\documentclass[preprint,12pt]{elsarticle}
\usepackage{color,soul}

\renewcommand\hl[1]{#1}  
\newcommand\updatedfigure{}
\newenvironment{updatedsection}{}{}

\let\today\relax
\makeatletter
\def\ps@pprintTitle{%
    \let\@oddhead\@empty
    \let\@evenhead\@empty
    \def\@oddfoot{\footnotesize\itshape
         {} \hfill\today}%
    \let\@evenfoot\@oddfoot
    }
\makeatother

\usepackage{amsmath}

\usepackage{tikz}
\usepackage{tikz-dependency}
\usepackage{multirow}%
\usepackage{manyfoot}%
\usepackage{booktabs}%
\usepackage{hyperref}
\hypersetup{pdfborder=0 0 0}
\usepackage{soul}

\usepackage{comment}
\usepackage{url}

\usepackage{breakurl}

\usetikzlibrary{trees,positioning,shapes,shadows,arrows}

\usepackage{amssymb}

\usepackage{natbib}

\newcommand\FM{FM}
\newcommand\GridFM{GridFM}
\newcommand\GridFMPF{GridFM\textendash v0}

\begin{document}

\begin{frontmatter}



\title{Foundation Models for the Electric Power Grid}

\author[ibmusa] {Hendrik F. Hamann\textsuperscript{\S*}}

\author[eth]{Blazhe Gjorgiev\textsuperscript{*}}
\author[ibmzrl]{Thomas Brunschwiler\textsuperscript{*}}
\author[eq]{Leonardo S. A. Martins\textsuperscript{*}}
\author[ibmzrl,epfl]{Alban Puech\textsuperscript{*}}
\author[eth]{Anna Varbella\textsuperscript{*}}
\author[ibmzrl]{Jonas Weiss\textsuperscript{*}}

\author[ibmdub]{Juan Bernabe-Moreno\textsuperscript{\dag}}
\author[hq]{Alexandre Blondin~Massé\textsuperscript{\dag}}
\author[nrel]{Seong Choi\textsuperscript{\dag}}
\author[anl,uchicago]{Ian Foster\textsuperscript{\dag}}
\author[cu,nrel]{Bri-Mathias Hodge\textsuperscript{\dag,}}
\author[nrel]{Rishabh Jain\textsuperscript{\dag}}
\author[anl]{Kibaek Kim\textsuperscript{\dag}}
\author[hq]{Vincent Mai\textsuperscript{\dag}}
\author[hq]{François Mirallès\textsuperscript{\dag}}
\author[hq]{Martin De~Montigny\textsuperscript{\dag}}
\author[hq]{Octavio Ramos-Leaños\textsuperscript{\dag}}
\author[hq]{Hussein Suprême\textsuperscript{\dag}}
\author[harvard]{Le Xie\textsuperscript{\dag}}
\author[hq]{El-Nasser~S. Youssef\textsuperscript{\dag}}
\author[hq]{Arnaud Zinflou\textsuperscript{\dag}}

\author[bra]{Alexander Belyi\textsuperscript{\ddag}}
\author[inesc]{Ricardo J. Bessa\textsuperscript{\ddag}}
\author[doe]{Bishnu Prasad Bhattarai\textsuperscript{\ddag}}
\author[ibmusa]{Johannes Schmude\textsuperscript{\ddag}}
\author[nyu,bra]{Stanislav Sobolevsky\textsuperscript{\ddag}}

\affiliation[ibmusa]{organization={IBM Research},
            addressline={T.J. Watson Research Center}, 
            city={Yorktown Heights},
            postcode={10598}, 
            state={NY},
            country={USA}}

\affiliation[ibmzrl]{organization={IBM Research - Europe},
            addressline={Säumerstrasse 4}, 
            city={Rüschlikon},
            postcode={8803}, 
            country={Switzerland}}

\affiliation[eq]{organization={Equilibrium Energy, Inc.},
            addressline={333 Kearny Street}, 
            city={San Francisco},
            postcode={94108}, 
            state={CA},
            country={USA}}

\affiliation[nrel]{organization={National Renewable Energy Laboratory},
            addressline={15013 Denver W. Parkway}, 
            city={Golden},
            postcode={80401}, 
            state={CO},
            country={USA}}

\affiliation[eth]{organization={Reliability and Risk Engineering Lab, Institute of Energy and Process Engineering, Department of Mechanical and Process Engineering},
            addressline={ETH Zürich}, 
            city={Zürich},
            postcode={8092}, 
            country={Switzerland}}

\affiliation[epfl]{organization={École polytechnique fédérale de Lausanne (EPFL)},
            addressline={Rte Cantonale}, 
            city={Lausanne},
            postcode={1015}, 
            country={Switzerland}}

\affiliation[anl]{organization={Argonne National Laboratory},
            addressline={9700 S Cass Av}, 
            city={Lemont},
            postcode={60439}, 
            state={IL},
            country={USA}}
            
\affiliation[doe]{organization={Department of Energy},
            addressline={1000 Independence Ave., S.W.}, 
            city={Washington, DC},
            postcode={20585}, 
            state={DC},
            country={USA}}

\affiliation[inesc]{organization={INESC TEC},
            addressline={Campus da FEUP, Rua Dr Roberto Frias}, 
            city={Porto},
            postcode={4200-465}, 
            country={Portugal}}

\affiliation[harvard]{organization={Harvard University},
            addressline={150 Western Ave}, 
            city={Allston},
            postcode={02134}, 
            state={MA},
            country={USA}}

\affiliation[ibmdub]{organization={IBM Research - Europe},
            addressline={IBM Technology Campus Damastown Industrial Park Mulhuddart Co}, 
            city={Dublin},
            postcode={D15HN66}, 
            country={Ireland}}

\affiliation[hq]{organization={Hydro-Québec},
            addressline={1800 Bd Lionel-Boulet}, 
            city={Varennes},
            postcode={J3X1S1}, 
            state={QC},
            country={Canada}}

\affiliation[uchicago]{organization={University of Chicago},
            addressline={5730 S Ellis Ave}, 
            city={Chicago},
            postcode={60637}, 
            state={IL},
            country={USA}}

\affiliation[cu]{organization={University of Colorado},
            addressline={1111 Engineering Dr}, 
            city={Boulder},
            postcode={80309}, 
            state={CO},
            country={USA}}
            
\affiliation[nyu]{organization={New York University},
            addressline={370 Jay Str}, 
            city={Brooklyn},
            postcode={11201}, 
            state={NY},
            country={USA}}
            
\affiliation[bra]{organization={Masaryk University},
            addressline={Kotlarska, 2}, 
            city={Brno},
            postcode={62000}, 
            country={Czech Republic}}




\end{frontmatter}

\noindent{\S~Lead contact: hehamann.hh@gmail.com}\\
{*~These authors contributed equally to this work as main contributors.}\\
{\dag~These authors contributed equally to this work as significant contributors.}\\
{\ddag~These authors contributed equally to this work.}

\section*{Keywords}
Foundation Models, Data-Driven Power Grid Modeling, Energy Transition,
AI-based Power Flow Simulation

\section*{Summary}
    Foundation Models (FMs) currently dominate news headlines. They employ advanced deep learning architectures to extract structural information autonomously from vast data-sets through self-supervision. The resulting rich representations of complex systems and dynamics can be applied to many downstream applications. Therefore, advances in \FM{}s can find uses in electric power grids, challenged by the energy transition and climate change. This paper calls for the development of \FM{}s for electric grids. We highlight their strengths and weaknesses amidst the challenges of a changing grid. It is argued that \FM{}s learning from diverse grid data \hl{and topologies}, that we call \GridFM{}s, could unlock transformative capabilities, pioneering a new approach in leveraging AI to redefine how we manage complexity and uncertainty in the electric grid. 
    \hl{Finally, we discuss a practical implementation pathway and road map of a \GridFMPF{}, a first \GridFM{} for power flow applications based on graph neural networks, and explore how various downstream use cases will benefit from this model and future \GridFM{}s.}

\section{Introduction}
\label{sec:introduction}
Power grids are arguably the largest machines ever built by humanity, as recognized by the National Academy of Engineering~\cite{constable2003century}. Having survived the deregulation revolution at the dawn of the new century, this aging and ever-expanding system is about to face its biggest challenge. The energy transition, with its widespread electrification and massive deployment of renewable and distributed generation, is the main driver for this change, affecting both the demand and supply sides of the market~\cite{MallapragadaDecarb2023,XinMarket2023,NijsseSolar2023}. And it is happening fast. Climate change, manifesting with more frequent high-intensity weather events, poses additional challenges to system planners and operators~\cite{Stankovski2023,LevinTex2022, Afsharinejad2021}.

\begin{figure}
    \centering
    \includegraphics[width=0.65\linewidth]{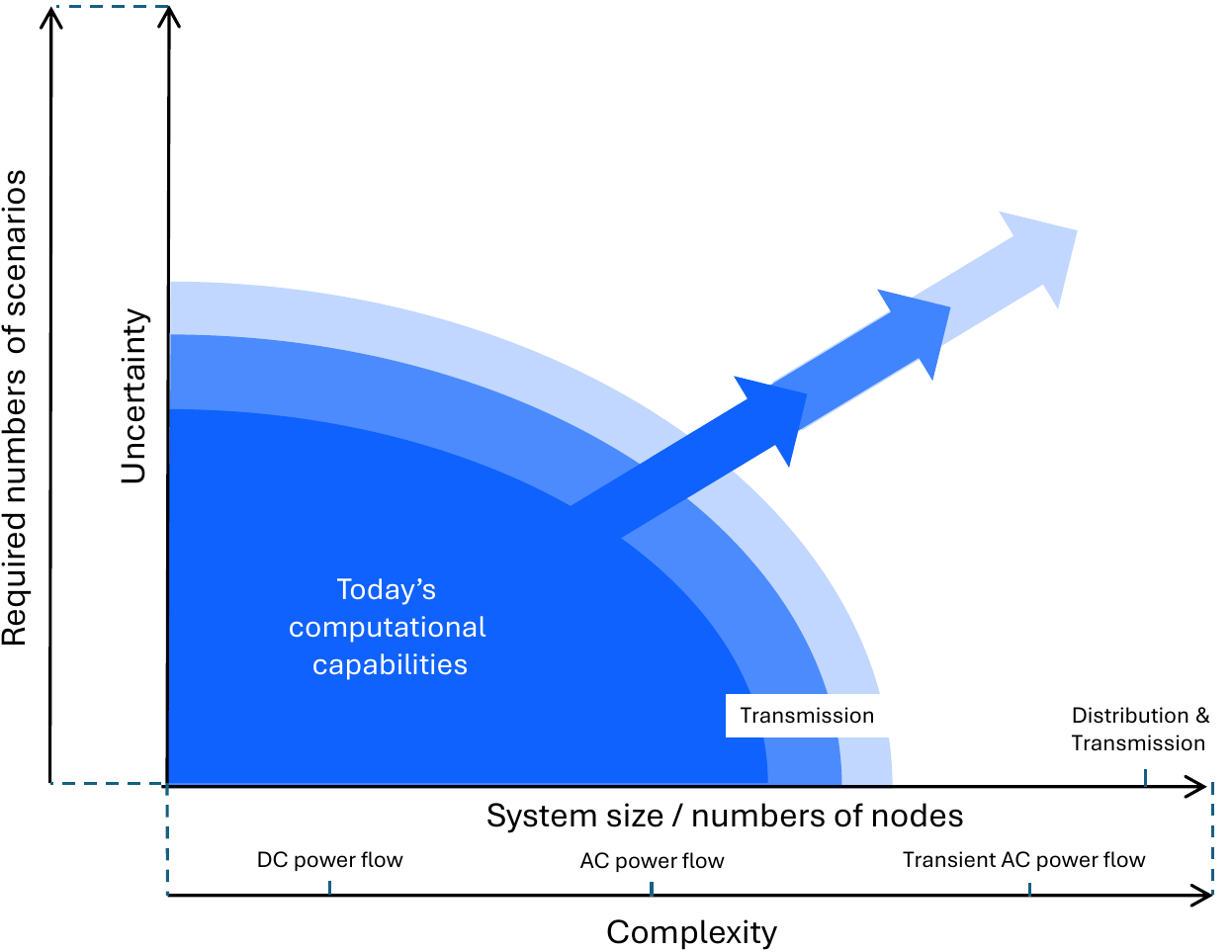}
    \caption{\updatedfigure The energy transition, aging infrastructure, cybersecurity challenges, and climate change greatly increase complexity and uncertainty in operating, controlling, and planning the power grid. This is creating a widening gap between existing computational capabilities and the evolving needs of the electric power industry.
    }
    \label{fig:energy-transition}
\end{figure}

In this context, it is clear that the operation, control, and planning of power systems will soon be pushed to their limits. As depicted in Figure~\ref{fig:energy-transition}, new computational methods and approaches are needed, capable of better tackling the challenges presented by increased uncertainty and complexity. Machine Learning (ML) and Artificial Intelligence (AI) methods have shown promise for such purposes in a wide spectrum of industries, with significant breakthroughs in computer vision~\cite{LeCunVision1989}, natural language processing~\cite{googleBert2019}, and intelligent control~\cite{SilverAlpgaGo2016}. Such approaches rely on statistical algorithms to learn from massive data, often enabling them to execute tasks (e.g., regression, classification, detection, segmentation, forecasting, and generation) without human supervision.

Yet, despite significant progress in the application of AI/ML methods to power systems~\cite{HarrisBatML2023,Gjorgiev2022,varbella2024powergraph}, wider adoption has been impeded, in particular, by a lack of readily available training data and the limited transferability of developed models to adjacent applications and \hl{system configurations}~\cite{MajumderPowerLLM2024}. This is where, we argue, Foundation Models (FM) have an important role to play. \FM{}s are advanced AI models developed through self-supervised learning, most often based on transformer architectures, that generalize across various tasks after initial training on large data-sets, enabling efficient adaptation to specific applications with minimal labeled data.
Text-based \FM{}s like GPT-4~\cite{chatGPT32020}, encoding deep relationships in text, have enabled a wide range of new applications. Similarly, we anticipate that \FM{}s for power grids, trained on the diverse data that characterizes the power grid, will enable a broad set of new use cases.  

In this article, we advocate for developing \FM{}s for power grids: what we refer to as \GridFM{}s. We explain why we believe in their potential, highlighting their anticipated strengths and weaknesses in light of the challenges posed by the evolving power grid. \hl{We introduce a practical road map for implementing \GridFMPF{}, an initial version of a \GridFM{} for power flow applications, and discuss our vision for future \GridFM{}s}, pioneering a new approach in leveraging AI to manage complexity and uncertainty in the power grid.

In the following, we first highlight, in Section~\ref{sec:challenges}, the looming power grid challenges associated with the energy transition, which are further aggravated by aging infrastructure, swelling cybersecurity, and climate change concerns. We show how, by increasing the complexity and uncertainty of the power grid, these drivers are creating a significant gap in computational capabilities, which we propose to overcome with \FM{}s. We then review in Section~\ref{sec:fm} the strengths and weaknesses of \FM{}s and propose in Section~\ref{sec:fm4grid} \hl{the near-term and long-term development phases of \GridFM{}s that we argue will overcome current limitations and enable scenarios, which are currently unattainable.} 

\bigskip
\section{Looming challenges}
\label{sec:challenges}

As the energy transition accelerates, power grids face unprecedented challenges threatening their reliability and efficiency.

\subsection{Distributed and renewable energy sources}
\label{sec:challenges:resources}
While a single coal plant can generate hundreds of megawatts predictably and reliably, large numbers of weather-dependent \textit{renewable energy resources} like wind and solar are required to produce the same power output. Current methods to forecast \hl{1-day or 1-hour} ahead still show up to 3\% to 10\% forecast error, respectively~\cite{Wang2022}, \hl{depending on the energy source.}
Additionally, Distributed Energy Resources (DERs) like battery storage and electric vehicles are operated by independent power management systems and are driven by user patterns with intrinsic uncertainties. 
%
%
Dealing with the added uncertainty from reverse DER power flows, multipoint injections, and weather-induced fluctuations requires new voltage regulation methods, protection schemes, and advanced (optimal) power flow analysis. The numbers of controllable and dispatchable devices, as well as balancing nodes are expected to grow exponentially~\cite{DallAnese2014DecentralizedSystems_dup} with additional thousands to millions to be added in the near future. Managing the associated variability and uncertainty will require advanced stochastic optimization, scenario reduction, and sampling-based approaches, which are computationally more intensive than currently used deterministic methods, especially for real-time demand-supply balancing.
Furthermore, as the net load of DER-rich distribution branches seen by transmission operators becomes more variable and less predictable, effective grid management and planning will become impossible without effective co-simulation frameworks that integrate transmission and distribution analysis. Yet the computational cost of such simulations, at least with existing technologies, is extremely high.

\subsection{Everything inverter}
\label{sec:challenges:inverter}
 
For the future power grid,\textit{ inverters} will play a crucial role in maintaining low cost at the grid's edge. However, with increased inverter proliferation, established control schemes and grid stability are challenged, while a more deregulated environment is created. 
%
With more inverters connected, grid inertia -- which stabilizes phase, voltage, and frequency during transient events and which is mostly tied to large synchronous generators -- decreases~\cite{ercot2015dc}, unless more grid-forming converters are introduced. Because inverters are controlled by software, they can quickly adjust power or disconnect, causing additional issues like sub-synchronous resonances and even trip large generators~\cite{Tan_2023}.

Despite the increasing adoption of standards~\cite{Kroposki_2023}, inverters remain hard to predict and model, unlike physics-driven synchronous generators. 
Quasi-static phasor simulations are used to assess optimal power flow. For transient grid stability assessment, however, computationally expensive Electro-Magnetic Transient (EMT) simulations~\cite{Choi_2021} are required. 
They run mostly on traditional Central Processing Units (CPUs), which are flexible for many algorithms but inferior in compute performance to recent Graphics Processing Units (GPUs), which are just starting to get adopted in the industry, as, e.g., in~\cite{Sun2021}. \hl{Also, CPU systems do not always scale well for larger EMT simulations when run on multiple CPU cores~\cite{paraemt1}. For example, consider ParaEMT -- NREL's EMT simulator~\cite{paraemt1, paraemt2} -- on a single CPU, runs a 1-second, 50 $\mu$s resolution EMT simulation for the WECC 240-bus system in 29 seconds. Another speed-optimized algorithm~\cite{Choi_2021} takes 11 minutes to compute a 0.25-second window for a 52-bus system at 1$\mu$s resolution.} Grid operations, however, would require EMT simulations to run in real-time and for large sections of the grid, including distribution grid sections. 

\hl{To further account for inherent network uncertainties and those introduced by Inverter-Based and Distributed Energy Resources (IBRs, DERs), stochastic ensembles of the problems must be simulated. Altogether, such an approach is computationally very expensive and, with current approaches, impractical~\cite{opal:2024}.} Thus, the community mostly relies on offline simulators like PS CAD~\cite{PSCAD2024} or ParaEMT~\cite{paraemt1} while starting exploring physics-informed neural networks (PINNs) to reduce EMT simulation time~\cite{Nellikkath2024}.

\subsection{Changes to demand and weather patterns}
\label{sec:challenges:demand}

Mass electrification of heating, transportation, and industry -- in addition to climate change and shifting demographics -- result in changing load profiles. Consequently, the accuracy of classical short-term load forecasting has declined over the past decade due, among other factors, to the changes in customer \textit{demand and weather patterns}, leading to out-of-distribution scenarios~\cite{hong2016probabilistic}.
To model demand uncertainty, utilities use multiple scenarios for structural and random drivers \cite{AEMO_2023}. Increasingly, they also forecast at the power transformer level to manage local dynamics \cite{Kampezidou_short_forecasting}, but this approach is costly and does not scale due to the need for multiple models. 

\paragraph{Short-term models} Short-term models for operations, running hourly to weekly, are used for market bidding, unit commitment/economic dispatch, operational planning, and security analysis, among others~\cite{Campillo2012EnergyDM}.
As stated before, uncertainty and variability are driven in part by weather and human behavior, from societal aggregations to individual households~\cite{Lazzari_2022}, as well as by operator-induced industrial activities. Economic factors, the increasing frequency and intensity of extreme weather events with high prediction uncertainty~\cite{Sheshadri2021}, and growing system complexity (e.g., self-consumption, electric vehicles) also contribute to variability. The resulting complexity necessitates the selection and integration of new and heterogeneous data sources (e.g., social networks, online news) along with the frequent adjustment, rebuilding, and maintenance of models.

\paragraph{Long-term models} Long-term models like the Regional Energy Deployment System~\cite{ReEDS_2024} are used for resource, infrastructure, and strategic planning. Uncertainty and variability at longer time scales are significantly driven by climate policy, technology adoption, and regulatory changes. Instances of this include the combustion engine phase-out by 2035 in Europe -- which resulted in a massive increase in the long-term electricity demand forecast -- or the US Federal Energy Regulatory Commission's rules on planning and covering the cost of the power grid~\cite{FERC_2024}. To improve model accuracy, additional information that captures demographic changes and changes in the distribution of wealth, which lead to different societal and individual behaviors (e.g., heating, cooling, electromobility), as well as climate change-induced environmental shifts must be integrated.
Therefore, the capacity to combine heterogeneous data sources, including information in textual format, is a fundamental requirement to improve predictability. 

\subsection{Aging infrastructure}
\label{sec:challenges:aging}
Managing the vast amount of diverse equipment and \textit{aging infrastructure} -- the US power grid has hundreds of thousands of miles of transmission and millions of miles of distribution lines with over 55,000 substations -- constitutes a major challenge on its own. While some lines last up to 80–100 years~\cite{xcel2021overhead}, insulators, transformers, and generators require frequent inspections, ideally through permanent sensing, which, given their age and approaching end-of-life~\cite{eia2024planned}, may not always be installed. Modeling these components in mostly unknown conditions is difficult and may rely on computationally expensive stochastic and ensemble models. Driven by growing demand, utilities are increasingly exploiting their aging infrastructure towards its operational limits \cite{ADEKUNLE2024}, reducing margins and leading to larger size contingency analysis problems \cite{Kaplunovich2016}.
In response, much faster grid analysis functions are required, e.g., including grid contingency events.

Moreover, the lowering of grid inertia through IBR integration (see Section \ref{sec:challenges:inverter}) is further amplified by the retirement of old synchronous generators. The more frequent use of synchronous generators for regulating purpose bring them more often in operating zones where mechanical or thermal stresses are significant, and thus accelerate the aging of the generators. IBR related switching transients also disrupt existing protection relay schemes\hl{~\cite{Alkhazim2022}} and increase grid behavioral uncertainty.

\subsection{Cybersecurity threats}
\label{sec:challenges:cybersecurity}
DER and IBR technologies are currently being developed and deployed without robust regulatory frameworks mandating essential cybersecurity standards, hence increasing the associated \textit{cybersecurity threats.} Interconnection and interoperability standards, such as IEEE Std.~1547 requires DERs and IBRs to communicate with clients, vendors, aggregators, and grid operators for monitoring and control. In practice, however, communication often occurs over inadequately secured public and private networks, expanding the cyberattack surface~\cite{Sandia2017}. 
Consequently, critical vulnerabilities are frequently discovered in commercial DERs and IBRs, such as electric vehicle charging stations~\cite{IEEE-TNSM-2024} and PV inverters~\cite{PVSC2017}, increasing the likelihood of major cyberattacks~\cite{IEEE-J-IoT-2022}. 
As the penetration of DERs and IBRs increases, synchronized cyberattacks targeting DERs or IBRs could cause grid reliability issues, from localized instability to system collapse~\cite{Sandia2017}. This risk is exacerbated by reduced grid inertia and contingency reserves due to reliance on zero marginal cost renewables. 
Additionally, control and management responsibilities for DERs and IBRs are often delegated to aggregators, vendors, and customers. This delegation complicates risk assessment, incident detection, and response -- as visibility barriers are created for grid operators -- introducing uncertainty and black-box behavior for large sections of their grid. With the lack of real-time data, these barriers create significant additional computational burdens for automated and accelerated data analysis to search for fraudulent injections.

\section{Foundation models}
\label{sec:fm}

\begin{figure}
    \centering
    \includegraphics[width=\textwidth]{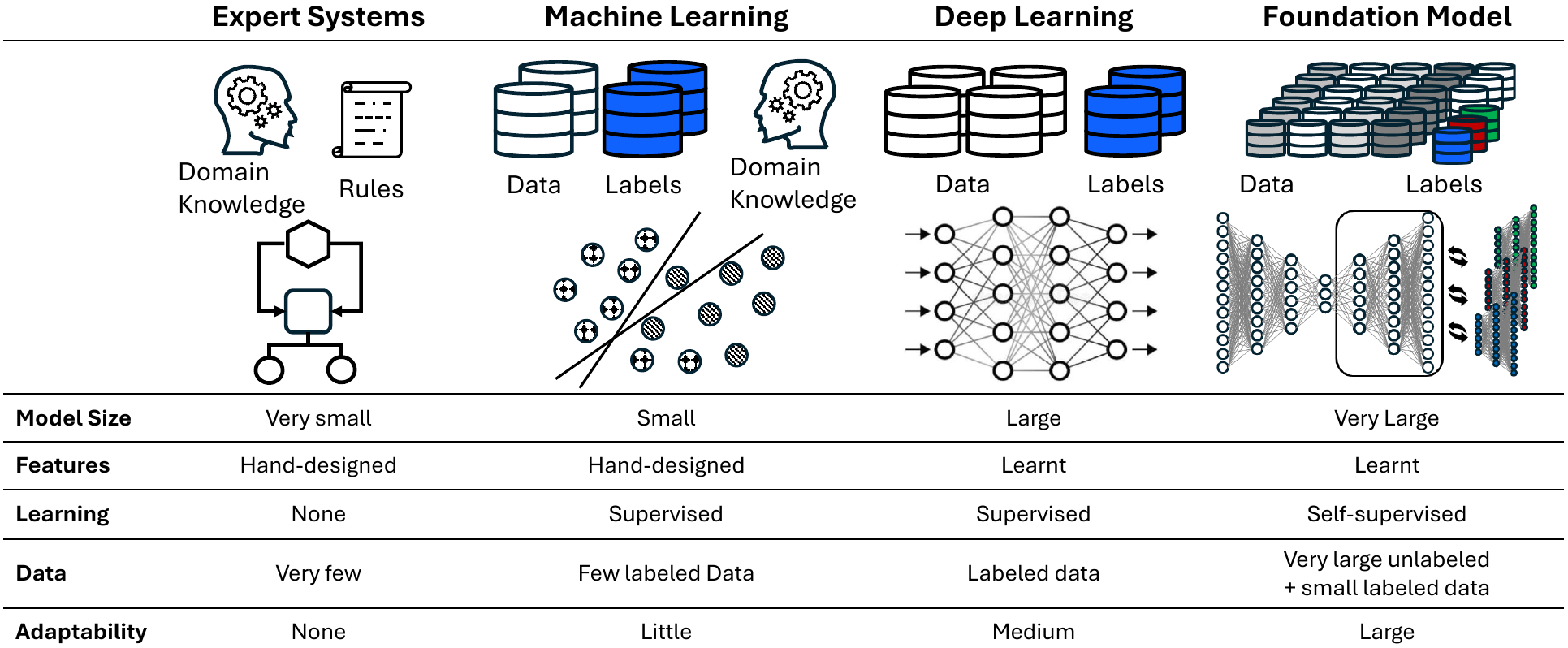}
    \caption{Evolution in AI and ML methods, leading to the emergence of \FM{}s}
    \label{fig:AI history}
\end{figure}

Having identified the growing gaps between computational capabilities and industry needs amidst the energy transition, we now introduce \FM{}s, laying the groundwork for our vision of \GridFM{}s.

\subsection{The foundation model approach}
\begin{figure}
    \centering
    \includegraphics[width=0.8\linewidth]{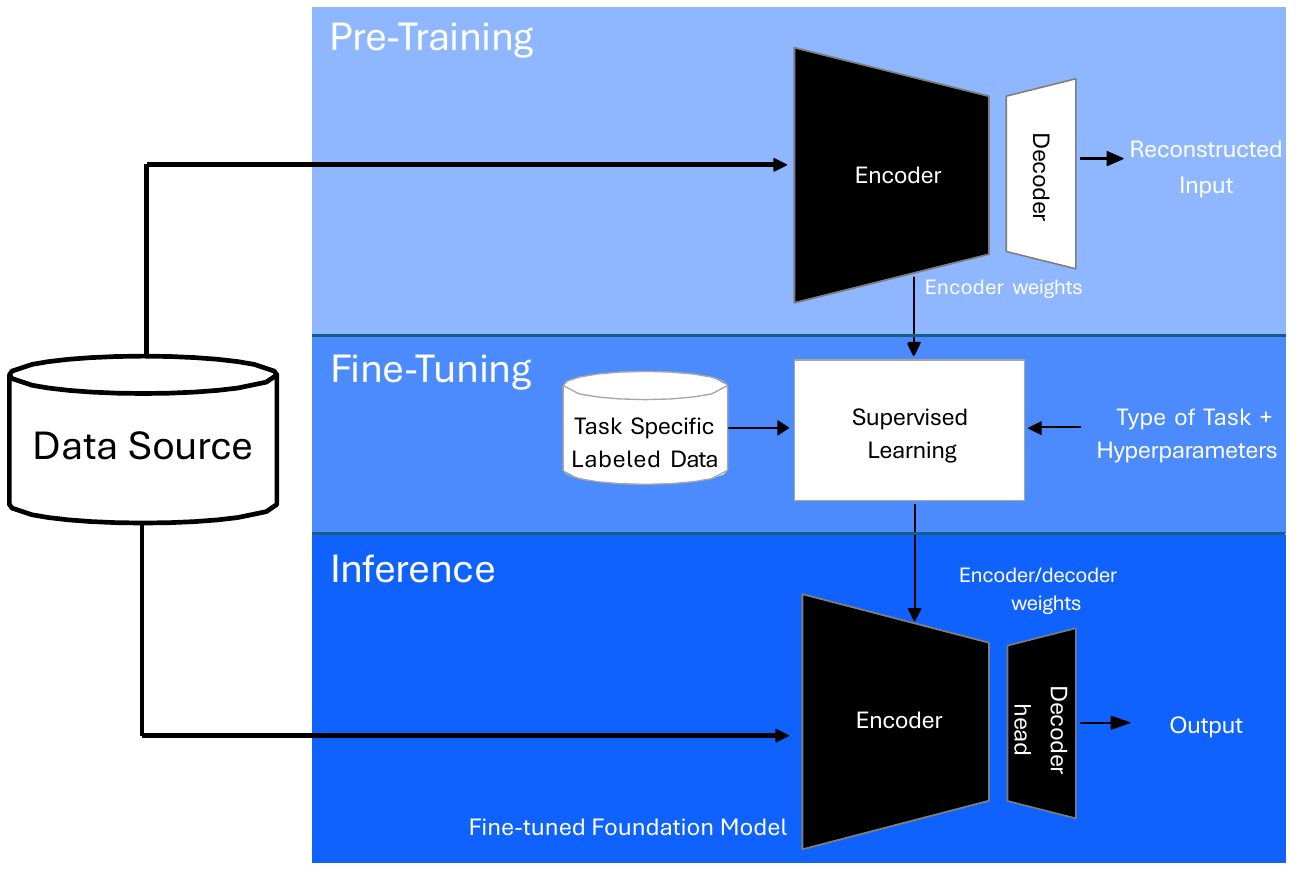}
    \caption{\FM{} life-cycle phases i) pre-training, ii) fine-tuning, iii) inference.}
    \label{fig:FMphases}
\end{figure}

 To understand the \FM{} approach, it is helpful to review the evolution of AI and ML as depicted in Figure~\ref{fig:AI history}. Early hand-crafted symbolic expert systems, from the 1960s onwards, were limited and brittle. The 90s brought big data and general-purpose ML that, however, \hl{still often} depend on task-specific hand-crafted features. In 2012, deep learning, enabled by increased compute performance, started to disrupt the AI world~\cite{krizhevsky2012imagenet}, but was \hl{more often than not used to train models on large annotated data-sets in a supervised way, making them task-specific}. Recently, \FM{}s have emerged. They learn from data through \hl{self-supervision and large content-rich data-sets}~\cite{bommasani2021opportunities} and have proven to generalize across many applications. \hl{\FM{}s draw their potential from being able to train on very large data-sets, which have not been annotated, thus are more easily available than labeled data-sets, as required for supervised learning.} The \FM{} approach is a three-step process, as shown in Figure~\ref{fig:FMphases}:

\begin{enumerate}
    \item \textit{Pre-training}\/: A general-purpose \FM{} is developed through self-supervised pre-training and a suitable architecture, often based on a transformer model with encoder and decoder components. This step involves a reconstruction task where the model reconstructs masked parts of the data.
    \item \textit{Fine-tuning}\/: The pre-trained \FM{} is customized for various downstream tasks and applications with minimal labeled data by attaching a task-specific decoder head and performing a few training iterations.
    \item \textit{Inference}\/: The fine-tuned \FM{} is deployed to allow users to request predictions at low computational cost.
\end{enumerate}

\hl{The development of \FM{}s involves extensive pre-training on data-sets of tens of terabytes and models with hundreds of millions or even billions of parameters~\cite{zhao2023survey}. However, in contrast to what one might assume, even bespoke AI models can be very complex and expensive to train. For example, the non-foundation and application-specific weather AI models, Graphcast (for forecasting) or CorrDiff (for downscaling) were trained on 32 TPU devices~\cite{lam2023graphcast} and on 128 A100 GPUs~\cite{mardani2023generative}, respectively.}

\hl{The key advantage of \FM{}s is that once developed, they can be readily adapted to multiple downstream applications via fine-tuning with minimal labeled data.
Thus, the cost of model pre-training can be spread across numerous applications. These economies of scale make it feasible to employ AI for niche applications, for which developing a dedicated traditional AI model might be prohibitively expensive or technically infeasible.} 

\FM{} pre-training encodes underlying patterns from the supplied data in model parameters. Successful examples of \FM{}s to date include text-based Large Language Models (LLMs)~\cite{zhao2023survey} but also weather forecast \hl{models~\cite{schmude2024prithviwxcfoundationmodel}}, in which these patterns are linguistic and physics-driven temporal-spatial relationships, respectively. LLMs have been shown to be applied successfully to certain power system problems \cite{MajumderPowerLLM2024}. By contrast, our focus here is on \hl{\GridFM{}s that go beyond text, that will eventually be able to handle multiple data modalities intrinsic to power grid operations.} Modalities of interest include network topologies, sensor data from 
Phasor Measurement Units (PMUs), \hl{alarms}, SCADA systems, Advanced Metering Infrastructure (AMI), load, generation, weather, and market data.

\subsection{Salient advantages}

We next review \textit{salient features} of \hl{AI in general and \FM{}s in particular} that hold significant potential for enhancing power grid applications.

\paragraph{Predicting ``tokens''} \label{Predicting tokens}
In transformer models, a token is defined as the smallest fundamental unit of data (e.g., a character sequence or image patch). \FM{}s have been shown to excel at predicting missing tokens based on context, enabling tasks such as sentence completion, programming assistance, and question answering in the text domain~\cite{mishra2024granite}. This capability extends beyond text to fields like time-series, audio, or video, where \FM{}s predict next or missing values~\cite{zeng2023transformers}. 

\paragraph{Homogenization and scaling} \label{Homogenization and scaling}
\FM{}s represent a paradigm shift in AI, offering a scalable and standardized approach and making niche applications commercially viable. This \textit{AI-as-a-platform} approach promises to simplify operations by allowing a few \FM{}s to replace many specialized models. For example, a single geospatial \FM{} for earth observation has been fine-tuned for tasks like flood segmentation, land-use classification, and heat island regression~\cite{jakubik2023foundation}.

\paragraph{Accelerating simulations} \label{Accelerating simulations}
\hl{AI models in general, including \FM{}s can emulate physics-based simulations with impressive accuracy and efficiency, potentially combined with physics-informed machine learning~\cite{Nellikkath2024}.} The efficiency gains originate from AI models directly mapping inputs to outputs, unlike traditional physics-based simulations that solve partial differential equations iteratively. Recent AI models~\cite{fourcastnet2022} emulate numerical weather prediction models~\cite{ECMWF2024} with comparable forecasting skills but up to 4–5 orders of magnitude lower computational cost~\cite{bi2023} during inference. These game-changing advances improve the spatial and temporal resolution of forecasts and, combined with AI techniques such as diffusion-based models~\cite{Price2024}, facilitate the creation of large ensembles for uncertainty estimation. Another salient advantage of using AI for emulating simulations lies in the fact that no a-priori assumptions are required, but rather the AI model can learn the full physics directly from the data or observations. We argue that exactly these characteristics of \hl{AI in general and \FM{}s in particular} are crucial for bridging the computational gaps shown in Figure~\ref{fig:energy-transition}.

\subsection{Hurdles to implementation} \label{sec:Hurdles to implementation}
Here, we discuss \textit{potential hurdles} associated with implementing \FM{}s for the power grid while suggesting solutions.

\paragraph{Need for data} \FM{}s, like other deep learning models, are data hungry. 
For example, the Llama3 LLM, with 70 billion parameters, is trained on 15 trillion tokens of text~\cite{llama3}. By way of comparison, the power sector globally generates multiple times that volume daily \cite{7884486}. Considering the groundbreaking capabilities of large language models in open-ended systems where context and semantics are not consistently described by mathematical equations, we are confident about this technology's potential for the power grid. \hl{Diverse and high-quality data-sets required for training grid FMs will, though, still have to be assembled to cover the full spectrum of scenarios and situations.} \hl{Addressing transmission grid problems first, where data is available in a more standardized form, e.g., PMU or power flow data will help demonstrate the value of \FM{}s for the grid. This will create incentives for data at the distribution level, which currently is much more heterogeneous and of less quality~\cite{ZHOU2024110483}, to be collected, curated, and made available in suitable formats for use with AI and FMs at a larger scale.}

\paragraph{Data accessibility} Data accessibility is another hurdle to overcome. Grid data are heterogeneous and distributed, are owned by multiple entities, and have special security and privacy measures. A potential solution to this barrier includes privacy-preserving federated learning~\cite{kairouz2021advances} and secure enclave technologies~\cite{hunt2018chiron}. Such approaches may enable multiple data owners to train models collaboratively without sharing their local data with a central server, thus preserving privacy and complying with data protection regulations. Additional schemes like differential privacy and homomorphic encryption can be integrated into federated learning frameworks~\cite{ryu2022appfl}, ensuring that sensitive information remains protected during training. Secure enclaves, in contrast, create a trusted execution environment for secure processing of sensitive and proprietary data during \FM{} training. Moreover, simulated data from real and synthetic systems can complement and provide scenarios under various conditions, such as extreme weather conditions, component failures, and cascading failure events~\cite{varbella2024powergraph}.
\hl{As discussed further in Section \ref{sec:fm4grid}, \GridFM{}s might be developed by synthetic data, which are then updated and verified with proprietary data behind the user's firewall.} However, it is important to consider that electrical networks evolve for various reasons, such as network reinforcement, which changes topologies and operating conditions, or new business models (e.g., local energy communities), which change the operating profiles of assets connected to the network, requiring continual learning of the \FM{}.

\paragraph{Trust} Notwithstanding the scenario in which all necessary data, whether synthetic or real, can be cleared and collected for use, there still remains the age-old question of whether the data and, consequently, a data-driven model can be trusted. In general, power system operators prefer interpretable models over black-box models so that the root causes of choices by advanced tools can be better understood, and heuristic solutions can be developed. While the debate on the trustworthiness of AI will continue, there is evidence that AI models can be as performant as first-principle physics-based models, \hl{at least for specific applications~\cite{nguyen2023climax}. Performance alone however does not seem to be necessarily a sufficient condition for trust because even conventional solvers can have stability and convergence challenges~\cite{tostado2021solving}, which we have found ways to deal with through decades of usage.}

\hl{Referring back to the recent breakthroughs of AI emulators for weather, it was demonstrated that AI could indeed encode the actual dynamics of certain physical systems and generalize to unseen events in ways that allow it to have comparable performances to a physics-based, simulation-based model~\cite{hakim2024dynamical}. In some cases, e.g. for an AI power flow emulator, which is discussed in Section~\ref{sec:fm4grid}, the AI results can be immediately verified by using the physics equations. In other cases, where such ad-hoc verification is not possible, a consistent way is required to evaluate the model for out-of-sample scenarios. In the longer-term a set of standards must be developed as to what performances must be achieved for a given out-of sample scenario. Towards that end, ISO has already developed the first family of standards, e.g., ISO/IEC 24029-2~\cite{ISO24029}.} The European Union's recently published AI Act~\cite{AIact} defines two concepts to boost trust in AI-based systems and data sharing: a) testing and experimentation facilities that combine physical and virtual environments to evaluate and certify their latest AI-based software and hardware, and b) the data spaces concept to remove barriers to data sharing, such as interoperability, trust, and privacy, while keeping sovereignty and full control of data.
\paragraph{Malicious use} \FM{}s can be used as a sandbox to accelerate the development of adversarial attacks. For example, \FM{}s' potential for scalable data imputation, combined with their ability to predict grid operating conditions, could be exploited by adversaries to infer sensitive information about a target power system~\cite{Youssef2023IEEETSG}. Therefore, it is imperative to continuously assess the extent to which misusing such technology can augment malicious actors’ attack capabilities and to adapt cybersecurity defenses and policies in response~\cite{DOE2024Rep}. 

\label{sec:challenges:Opportunities}

\section{\hl{Proposed implementations for \GridFM{}}}
\begin{updatedsection}
\label{sec:fm4grid}

\begin{figure}[t]
\centering
\includegraphics[width=1\linewidth]{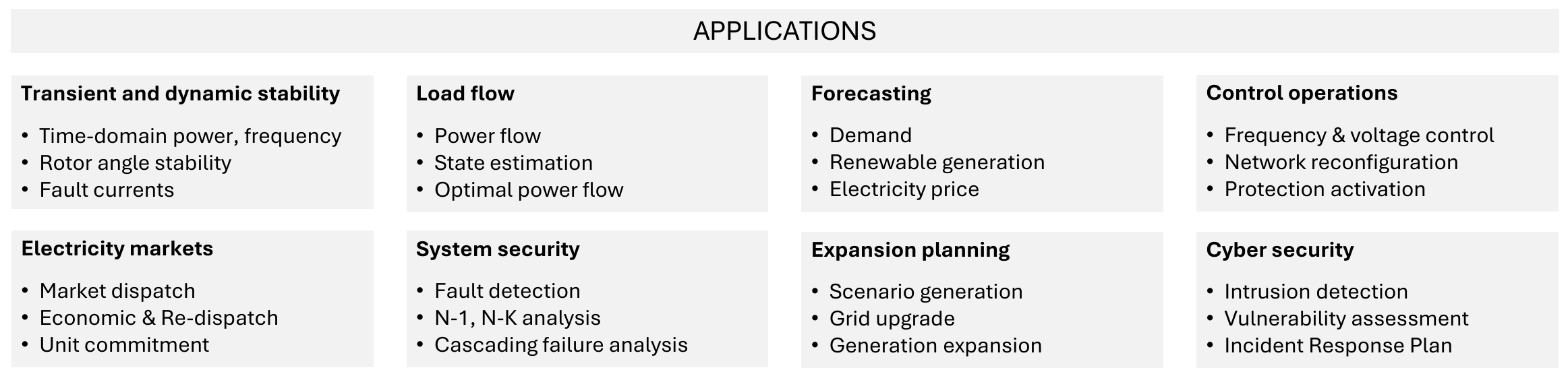}
\caption{Applications: Examples of power system problems to be solved with \GridFM{}s.}
\label{fig:downstreamstask}
\end{figure}
 
The broader vision of a \GridFM{} is the enablement of all or as many applications as possible. However, even if initially \GridFM{}s would only be suitable for a single application, the superior adaptability of FMs over bespoke AI models through fine-tuning to thousands of different utilities with their specific data and requirements justifies the FM approach.

In the above context, fine-tuning can mean different things. 
A \GridFM{} may be fine-tuned to a set of diverse applications like load forecasting, fault detection and others. Initially, this may be too ambitious and challenging, considering the diversity of downstream applications as shown in Figure~\ref{fig:downstreamstask}, each requiring different data, temporal scales, and pre-training strategies. 
But fine-tuning may also refer to tailoring a \GridFM{} for a given application with user-specific (and potentially private) data to the particular user needs and requirements. 
For example, just in the United States, there are 3,200 utilities~\cite{us_electricity_primer_2015}, with their own private and siloed data and grid-specific requirements. These utilities may not have sufficient data by themselves but could directly profit from the proposed approach. 
This multi-level fine-tuning approach is not unlike what can be observed in geospatial or weather foundation models, where FMs have shown excellent adaptability to different geographic regions, spatial and temporal resolutions for a given downstream application~\cite{keisler2022forecasting, mukkavilli2023, schmude2024prithviwxcfoundationmodel}. 

We thus propose an implementation road map as outlined in Figure~\ref{fig:road map}, with a first, more specific version of \GridFM{}s, pre-trained for power flow applications (\GridFMPF{}), which is expected to provide real-world benefits in the next 1-3 year. 
In the long-term, we envision expanding and addressing more grid applications by including more diverse data modalities to pre-train a single \GridFM{} or by combining different \GridFM{}s, as discussed in Section \ref{sec:long-term}.   
The rationale to start with power flow is as follows:

\paragraph{(1) A highly relevant pre-training task} \FM{}s are known to excel when the pre-training task is closely related to the enabled downstream applications~\cite{HAN2021225}. 
Consequently, with AC power flow at the core of power system analysis~\cite{powerflow}, its reconstruction is an ideal pre-training task for \GridFMPF{}. This enables power flow related applications in the area of load flow (optimal power flow, state estimation, power flow), system security (contingency analysis, cascading failure), and beyond.

\paragraph{(2) AI power flow models are computationally more efficient} Current power flow solvers, e.g. using Newton-Raphson methods, are computationally expensive, especially for large grids. For instance, for an N-1 contingency analysis on the Western Interconnection, it takes approximately 5 minutes per scenario~\cite{benes2024ai}, while meeting the NERC TOP-001-3 requires running these analyses within a limited time period.
In contrast, \GridFMPF{} is projected to achieve a computational speed-up of 3 to 4 orders of magnitude over conventional solvers, as demonstrated on IEEE grid cases with up to 118 buses~\cite{DONON2020106547}. For larger grids, we anticipate a similar or greater speedup, as GNNs exhibit linear complexity in the number of grid buses~\cite{Wu_2021}, while Newton-Raphson methods scale quadratically with the number of buses~\cite{4596138}. This significant efficiency gain will enable real-time analysis of multiple scenarios of more complex grids~\cite{DONON2020106547}, which will become crucial for maintaining system stability and optimizing future grids under uncertainties stemming from the energy transition and from climate change~\cite{benes2024ai}.

\paragraph{(3) There have been significant advances in GNN-based power flow}
Using GNNs to represent the power grid and mapping power flow to the message-passing scheme of GNNs has been successfully demonstrated recently~\cite{DONON2020106547}. These models replicated the results of traditional power flow solver with less than 1\% error on the predicted quantities, despite past GNN training challenges~\cite{NEURIPS2022_23ee05bf}. While these models were trained in a supervised way, recent results of the previously discussed GNN-based weather \FM{}s~\cite{keisler2022forecasting, mukkavilli2023, schmude2024prithviwxcfoundationmodel} indicate that learning power flow with GNNs in a self-supervised way is a valid approach for a \GridFM{}. Furthermore, large-scale data foundation models (FM) based on GNNs have already been developed~\cite{lam2023graphcast}, and a comprehensive review can be found in \cite{mao2024graph}.

\begin{figure}[t]
\centering
\includegraphics[width=1\linewidth]{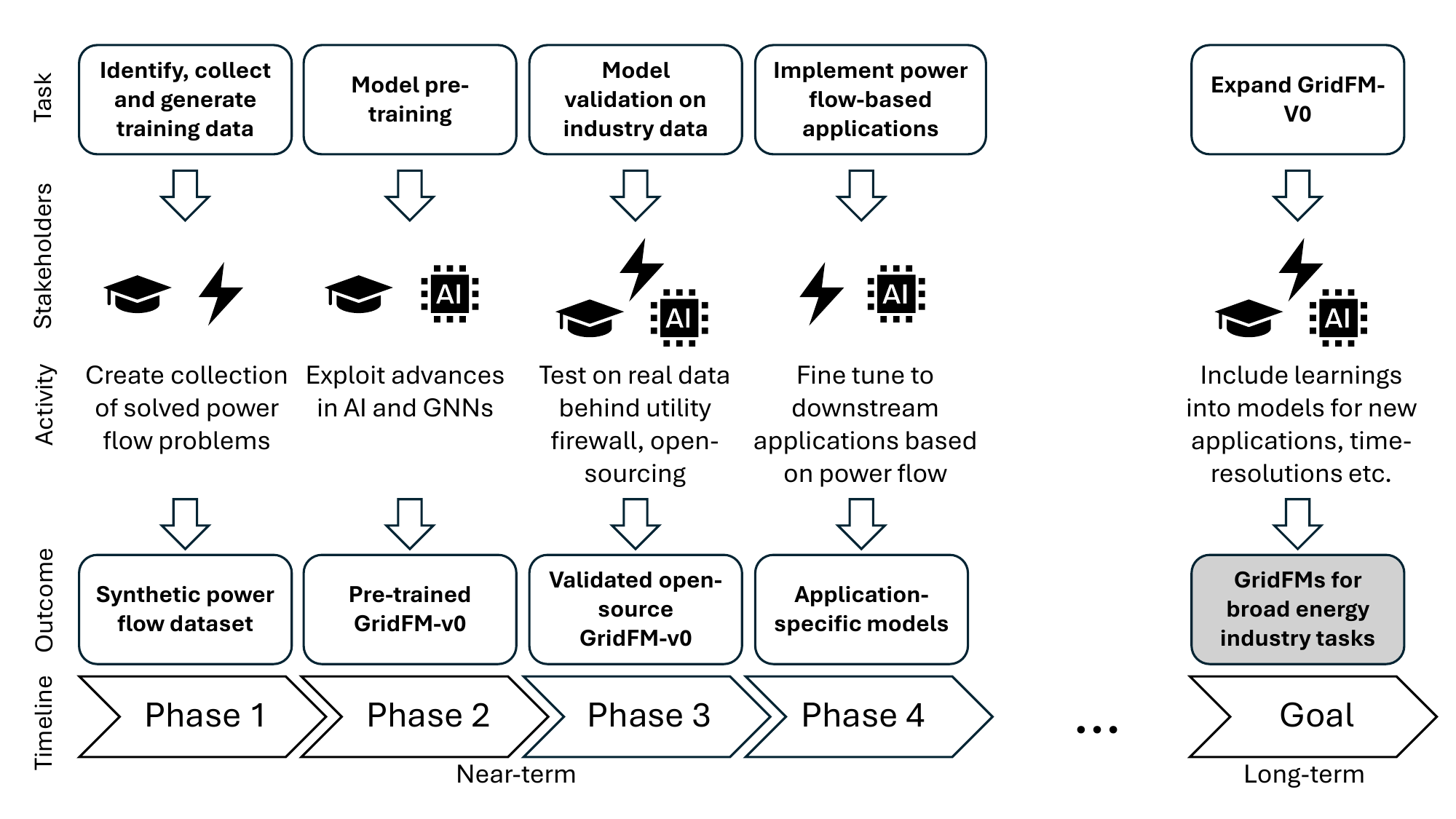}
\caption{GridFM implementation road map with four near-term phases to develop \GridFMPF{}, a \GridFM{} for power flow related applications. The icons refer to academia, power system industry, and AI community. The long-term goal is creating a family of \GridFM{}s for a broader range of grid challenges.}
\label{fig:road map}
\end{figure}

\subsection{\hl{Near-term: \GridFMPF{}, a \GridFM{} for power flow applications}}
\label{sec:short-term}

The four phases as foreseen for the development of \GridFMPF{}, and as depicted in Figure~\ref{fig:road map} are discussed in detail in the following.

\subsubsection{\hl{Phase 1: Identify, collect and generate training data}}

Training \GridFMPF{} requires a large collection of solved power flow problems for different grid topologies, parameters, and different load conditions. In recent years, large power grid data-sets have been made available, created by (1) collecting grid topology and measurement data, (2) generating varying synthetic load conditions per grid model and (3) using established solvers to compute a wide range of power flow solutions under different operational conditions.We aim at using the datas-et provided by~\cite{varbella2024powergraph}, which contains power flow solutions under real load conditions on 4 different grid topologies. Additionally, to complement this data, diverse power flow problems using traditional power flow solvers~\cite{pandapower, powermodels,matpower} will be computed, with power grid topologies from PGLIB-OPF~\cite{pglib} and standard IEEE benchmarks and with topology and load perturbations. Consequently, our model will be pre-trained on data from realistic topologies with a broad range of operational conditions. 
This approach facilitates the model to generalize well across topologies, as demonstrated in~\cite{VarbellaCFS2023}. 
For model fine-tuning, data-sets from~\cite{VarbellaCFS2023} for system security analyses and \cite{lovett2024opfdatalargescaledatasetsac} for Optimal Power Flow (OPF) will be used. The latter contains 300,000 solved optimal power flow problems for ten different grid topologies each, including solutions for N-1 contingency analysis, with grid sizes ranging from 14 to 14,000 buses. 

\subsubsection{\hl{Phase 2: Model architecture development}}

\begin{figure}[t]
\centering
\includegraphics[width=1\linewidth]{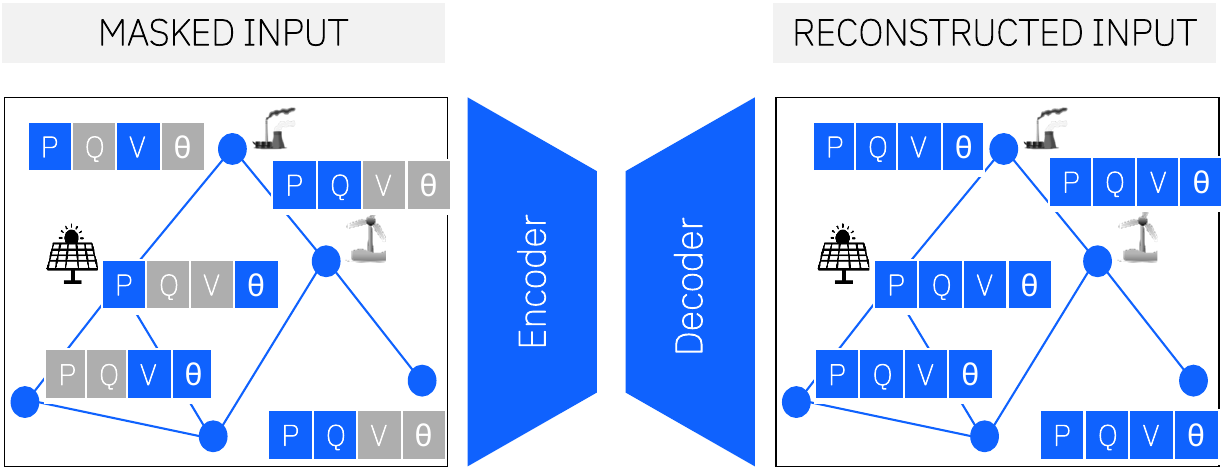}
\caption{\GridFMPF{} is pre-trained to reconstruct masked power flow data from input graphs.}
\label{fig:FMconceptnear}
\end{figure}

In the second phase, suitable architectures will be evaluated by pre-training performance analysis, and by adjusting models, training strategies, and loss functions.
Figure~\ref{fig:FMconceptnear} illustrates the Masked Auto-Encoder (MAE) pre-training task of \GridFMPF{}. Intuitively, the grid is modeled as a graph, where nodes represent electrical buses with active power, reactive power, voltage magnitude, and voltage angle, \( (p_i, q_i, v_i, \delta_i) \), as node variables. We opted for this graph representation of power systems, but others can also be considered~\cite{Hines2010}. Transmission lines and transformers are modeled as graph edges. An autoencoder \cite{graphmae, graphmae2} is trained in a self-supervised manner to reconstruct masked node features from power flow solutions, each corresponding to a specific topology and load condition. This amounts to solving the power flow equations.
During training, the loss function ensures meaningful reconstruction of node variables within the physical constraints. 
The total loss \( \mathcal{L} = \mathcal{L}_{SCE} + \mathcal{L}_{power flow} \) combines the Scaled Cosine Error (SCE) \cite{graphmae}, which minimizes the distance between the original and reconstructed graphs, and \( \mathcal{L}_{power flow} \), which ensures that the power flow equations are satisfied. Recent studies show that an approach using physics-informed methods is the most effective for solving OPF. For example,~\cite{Chatzos2021} demonstrated that a hybrid loss function provides high-fidelity approximations for large-scale OPF, while \cite{Nellikkath2022} confirmed that this approach best solves AC OPF by integrating data-driven learning with physical laws. \FM{}s, as a concept, are designed to incorporate diverse methods, including GNNs and PINNs, into a unified framework, enabling them to generalize effectively and adapt to various tasks, such as solving OPF problems.

\subsubsection{\hl{Phase 3: Model scaling and validation on industry data}}

Due to regulatory constraints, it is unlikely or at least challenging to gain full access to sensitive data from operational power grids \cite{chai2024defininggoodevaluationframework, nsf_doe_privacy_research,doe_airmp}. To ensure the model's accuracy and relevance in practical applications, it is therefore imperative that a \GridFM{} be validated on operational data and real grid topologies.
To address these challenges, an iterative approach is adopted. Our \GridFMPF{} will initially be pre-trained and tested on openly available data. Subsequently, it will be re-trained and validated in collaboration with electricity utility partners behind the firewall in a secure compute environment. The results of these validations will guide us to generate more suitable synthetic training data, update the architecture, loss functions etc. so that the improved \GridFM{} yields satisfactory results in real-world conditions once fine-tuned. This approach ensures that the \GridFM{} remains performant and aligned with industry needs without exposing nor pre-training the model on proprietary data.

\subsubsection{\hl{Phase 4: Implement power flow-based use cases}}

\GridFM{}s are anticipated to achieve a computational speed-up of 3 to 4 orders of magnitude over traditional power flow solvers, as stated previously, which will directly benefit various applications:

 \paragraph{(1) Complementing Numerical Power Flow Solvers}
For applications that require solving power flow frequently, and where speed is critical, linear approximations (such as DC power flow~\cite{7798623}) are typically used. For these applications, faster AI-based power flow solvers offer a trade-off between ``traditional'' DC power flow and AC power flow solutions. We expect minimal fine-tuning of these models, as the task is equivalent to the suggested pre-training reconstructing task.
                 
\paragraph{(2) Contingency Analysis} Contingency analysis is essential for assessing grid resilience. 
On a grid with 1000 lines, simulating the loss of each individual line requires solving 1,000 distinct power flow problems. However, when considering scenarios that involve the simultaneous loss of any two components, the number of power flow analyses increases dramatically to \(\binom{1000}{2} = 499,500 \). 
By utilizing a power flow solver that is 1000 times faster, operators will be able to perform contingency analyses across a much larger set of scenarios. For example, solving 499,500 scenarios with AI-based methods would take roughly the same time as solving only about 500 scenarios using conventional physics-based solvers. Therefore, a \GridFMPF{} will greatly enhance grid analyses and contribute to operating and planning under uncertainty. 

The power grid is continuously evolving, therefore, the \GridFMPF{} must be able to account for different network topologies and different operating conditions of a same grid. This can be accomplished by training the model on grids with topological perturbations, as already demonstrated by \cite{piloto2024canosfastscalableneural}, or by additionally training the same model on entirely different grid topologies, as done in \cite{VarbellaCFS2023}.
          
\paragraph{(3) Optimal Power Flow (OPF)} 
The authors in~\cite{piloto2024canosfastscalableneural} trained a model on synthetic data and demonstrated an OPF speedup of 2 to 4 orders of magnitude while maintaining errors of less than 1\% in the objective value.
Solving the OPF translates into fine-tuning the \GridFM{} for reconstructing the generator power to minimize generation costs. Reconstructed node variables must adhere to bounds (e.g., generator capacity constraints), making this problem challenging. However, \GridFMPF{} can be hierarchically integrated into optimization schemes where generator power is iteratively adjusted, solution feasibility is checked, and other variables are determined using the power flow solver. 
               
\subsection{\hl{Long-term Vision: \GridFM{}s for a broader set of applications in power systems}}
\label{sec:long-term}
In Section~\ref{sec:short-term}, we explored an initial version of \GridFM{} for power flow applications, already addressing important challenges of the industry. The long-term objective, however, is to address as many applications from Figure~\ref{fig:downstreamstask} as possible, using a single \GridFM{} or a family thereof, fully exploiting \FM{} capabilities. Here we discuss how to build new industry-relevant applications on top of \GridFMPF{}.

\subsubsection{\hl{Extending \GridFMPF{} to a \GridFM{} capable of handling new applications}}

\paragraph{Forecasting} 
\GridFMPF{} was pre-trained to reconstruct power flow data for a single time step. This can be readily extended to pre-training to reconstruct power data over multiple consecutive time steps, similar to Weather FMs \cite{mukkavilli2023}. This enhancement will enable the forecasting of future grid states. By incorporating weather and climate data into the node embeddings, we will further improve the model's ability to predict generation and anticipate weather-induced grid disturbances.

\paragraph{Transient and Dynamic Stability Analysis} In addition to training \GridFM{}s on static power flow data, like in the case of \GridFMPF{}, the pre-training can additionally involve reconstructing high-resolution real-time data obtained from PMUs, which enables effective transient state estimation. This approach will facilitate rapid fault current detection and localization, efficient system restoration, and disturbance mitigation~\cite{ChenSansa2018}. Additionally, the generator rotor angle and voltage differential-algebraic equations can be added to the model using physics-based machine learning techniques~\cite{PS_Pys-ML} to, e.g., assess the grid's stability margins.

\paragraph{Control operations} 
There are two concrete ways in which \GridFM{}s will help in control operations. First, the generative capabilities of \FM{}s as a \textit{conditional load and generation scenario generator}, can be used to construct scenario trees that are inputted to multistage stochastic programming algorithms, as done in \cite{Heitsch,1304379}. Second, we can use \GridFMPF{} and \GridFM{}s in general as digital twins of power grids within Reinforcement Learning (RL) environments. Leveraging digital twins (data-driven representations of systems) enables faster computation of state transitions during RL training. This also allows RL agents to be trained on systems with highly complex dynamics that are difficult to model explicitly, such as distribution networks. As demonstrated in \cite{HUA2023121128}, a digital twin of a distribution network was used to train an RL agent, improving decision-making in both planning and operations. Third, \GridFM{}s can also be used as a pre-trained representation for RL agents, for which \FM{}s have been shown to be a valuable asset \cite{Yang2023FoundationOpportunities}.

\paragraph{Expansion planning} The pre-training of \GridFMPF{}, which involves reconstructing node-level masked power flow data, can be complemented by edge masking and reconstruction. Extending this approach further, we will leverage the model's generative capabilities to also propose new grid topologies that optimize for capacity expansion. By integrating constraints such as network reliability, load growth forecasts, and economic cost factors, \GridFM{} can generate topologies that minimize congestion, maximize redundancy, and ensure efficient power distribution. Optimizing grid topology and resource dispatch co-optimization has already been achieved with GNNs in \cite{AUTHIER2024110817}.

\paragraph{Cybersecurity} 
\GridFM{}s can be fine-tuned for anomaly detection tasks to identify patterns that signal potential cyber intrusions or physical failures \cite{Halvorsen2024}. We can also use \GridFM{}s to identify vulnerabilities and conditions that lead to cascading failures \cite{He2024}.
\newline\newline
Although the use cases mentioned above are the most apparent, experience with \FM{}s and LLMs shows that once a model is available, numerous new applications can emerge. Therefore, the reader is encouraged to begin considering these possibilities, as \FM{}s for the power grid will soon become a reality.

\subsubsection{\hl{Building multi-modal \GridFM{}s}}

Ideally, \GridFM{}s should be able to handle more than one set of application tasks. This can be done by combining multiple \GridFM{}s together~\cite{Yang2023FoundationOpportunities} or by building self-contained stand-alone larger models, ingesting multi-modal spatial, temporal, as well as text data~\cite{MajumderPowerLLM2024}. Finally, concepts such as Mixture-of-Experts (MoE) frameworks~\cite{shazeer2017outrageously} can be exploited to compose \GridFM{}s from a family of \GridFM{}s. 

Regardless of the method used, \GridFM{}s will need to handle diverse data modalities in the long term, to account for complex external factors—such as climate, weather, economic and societal influences, distributed power sources, and the electrification of mobility and heating~\cite{benes2024ai}. Training multi-modal FMs requires creating tokens for each modality, aligning them in space and time, and potentially normalizing or reducing their dimensionality~\cite{Jain2021}. Consequently, tasks beyond power flow and grid topology reconstruction~\cite{linkpred} will need to be incorporated for pre-training. This approach has been demonstrated with models like 4M~\cite{mizrahi20234mmassivelymultimodalmasked}, which unifies different modalities into discrete tokens for computer vision. We anticipate that the number of trainable parameters for such models could approach those of weather models, reaching hundreds of millions~\cite{bi2023,nguyen2023scaling}.

\end{updatedsection}

\clearpage

\section{Conclusion and Discussion}

The rapid emergence of \FM{}s is reshaping the AI landscape. While \FM{}s are well established for text and image analysis, here we have introduced the idea of \GridFM{}s: \FM{}s for the power grid, able to learn from \hl{grid-relevant} data to improve power grid operations, planning, and control. \hl{We first highlighted the strengths of \FM{}s, which include excellent token prediction skills (signal reconstruction and forecasting), broad adaptability and scalability, and simulation acceleration.} 
\hl{To demonstrate that \GridFM{}s are well suited to close the growing gaps between computational capabilities and the industry’s need to cope with the increasing uncertainty and complexity of the power grid, we draw on the experiences of \FM{}s for weather forecasting. The latter is a domain characterized by apparently intractable complexity.} Nevertheless, we have seen scalable and adaptable \FM{}s to produce much more computationally efficient models by more than 4 orders of magnitude than traditional weather forecasting models.


\hl{We have outlined concrete implementation phases to create an initial version, \GridFMPF{}, which is pre-trained on power flow data and tailored to accelerate power flow-related downstream applications (e.g., optimal power flow and contingency analysis). In the same way that \FM{}s have been able to learn the fundamental equations governing weather, we envision \GridFMPF{} to learn power flow, thereby creating a much more efficient model.} The key advantage of such a \GridFMPF{} is that once pre-trained on multiple grids and various networks and operating conditions, it can be readily fine-tuned to specific grid topologies and \hl{user requirements based on their own private data.}


\hl{This paper advocates for research and development of \FM{}s for the power grid, to facilitate the energy transition through collaboration between the various stakeholders. All stakeholders will benefit from \GridFM{}s, simply by fine-tuning it to their specific needs and with their proprietary data and knowledge -- without re-developing a full bespoke AI model each time. Towards that end, we envision, under the leadership of industry associations and open source initiatives, for the community of stakeholders to join forces and make the required investments to research and develop \GridFM{}s in open source, thereby revolutionizing power grid analysis and leveraging AI advancements to meet common goals and needs.} 

\section*{Author contributions} 
\begin{itemize}
    \item H.F.H.: Conceptualization, Investigation, Writing -- Original Draft, Project Administration
    \item B.G., L.S.A.M., A.P., A.V., J.W.: Conceptualization, Investigation, Writing -- Original Draft
    \item J.B., A.B.M., T.B., S.C., I.F., B.H., R.J., K.K., V.M., F.M., M.D.M., O.R., H.S., L.X., E.S.Y., A.Z.: Investigation, Writing -- Original Draft
    \item A.J.B., R.J.B., B.P.B., J.S., S.S.: Writing -- Review \& Editing
\end{itemize}

\section*{Lead contact}


Requests for further information and resources should be directed to and will be 
fulfilled by the lead contact, Hendrik F. Hamann (hehamann.hh@gmail.com).

\section*{Acknowledgment}
The work was supported by the respective organizations of the authors. The authors acknowledge the invaluable discussions and interactions in the working group for Foundation Models for the Power Grid, including Roya Ahmadi, Rich Argentieri, Gianni Vittorio Armani, Venkat Balachandran, Ranjini Bangalore, Venkat Banunarayanan, Edmon Begoli, Larry Bekkedahl, Keith Benes, Michelle Berti, Budhu Bhaduri, Roberto Bianchessi, Gilbert Bindewald, Louise Blais, David Bromberg, Matthias Bucher, Eren Çam, Supriya Chinthavali, Julien Chosson, Joseph Collum, Magnus Cormack, Peter De Bock, Fei Ding, Kathy Duviella, Darlain Edeme, Yamshid Farhat, Mark Fenton, Marcus Freitag, Alirez Ghassemian, Tassos Golnas, Gavin Goodland, Luca Grella, John Grosh, Benois Grossin, Andy Hackett, Hagen Haentsch, Zeryai Hagos, Grahame Hay, Frank Hellmann, Andy Henton, Raffael Hilber, Dan Hilton, Gabriela Hug, Alex Hurley, Sorana Ionescu, Kevin Johnston, Gareth Jones, Adrian Kelly, Bernd Klockl, Teddy Kott, Benjamin Kroposki, Teja Kuruganti, Larry Light, Dalton Lunga, Matt Magill, Adrian Maldonado, Antoine Maret, Pete Marsh, Russel Mason, James McClean, John Messer, Nikki Militello, Jordan Murkin, Xabi Muruaga, Indu Nambiar, Christian Nauck, Caleb Northrop, Gabby Nyirjesy , Marco Padula, Patrick Panciatici, Raul Alfaro Pelico, Mark Petri, Pierre Pinson, Abhishek Pontis, Josh Power, Alan Pressman, Feng Qiu, Mark Rafferty, Pippa Robertson, Nicoletta Rocca, Caroline Rose-Newport, Lyndon Ruff, Bryan Sacks, Kelly Sanders, Giovanni Sansavini, Michael Sheppard, Robert Shorten, Vikas Singhvi, Pratik Sonthalia, Brian Spears, Declan Stock, Olle Sundstroem, Matthew Tarduogno, Roger Thompson, Anna Carolina Tortora, Alex Vivas, Lucy Vu, Ben Weiner, Casey Werth, Daniel Wetzel, Gary White, Xin Wu, Steven Wysmuller, Yunchi Yang, Thomas Zadlo, Marzia Zafar and many others. This material is based upon work supported by the U.S. Department of Energy, Office of Science, under contract number DE-AC02-06CH11357.

\section*{Declaration of Interests}
The authors declare no competing interests. H.F.H is also affiliated with Yamagata University and University of Illinois Urbana-Champaign. S.S. is also affiliated with Masaryk University.

\section*{Declaration of Generative AI and AI-Assisted Technologies}


During the preparation of this work, the authors used ChatGPT to find more concise reformulations and to condense paragraphs. After using this tool, the authors reviewed and edited the content as needed and take full responsibility for the content of the publication.





\bibliographystyle{elsarticle-num} 
\bibliography{
references-full.bib}





\end{document}